\documentclass{article}
\usepackage{graphicx}
\usepackage[round]{natbib}
\usepackage{xcolor}
\usepackage{hyperref}
\usepackage{authblk}
\usepackage[margin=3cm]{geometry}

\hypersetup{
    colorlinks,
    linkcolor={red!50!black},
    citecolor={blue!50!black},
    urlcolor={blue!80!black}
}

\title{The practice of qualitative parameterisation in the development of Bayesian networks}

\author[1,2]{\href{mailto:<steven.mascaro@monash.edu>?Subject=Qualitative Parameterisation}{Steven Mascaro}{}}
\author[1,2]{Owen Woodberry}
\author[3]{Yue Wu}
\author[1]{Ann E. Nicholson}

\affil[1]{
    Bayesian Intelligence\\
    Upwey, Melbourne, Australia
}
\affil[2]{%
    Faculty of Information Technology\\
    Monash University\\
    Clayton, Melbourne, Australia
}
\affil[3]{%
    School of Public Health\\
    University of Sydney\\
    Sydney, Australia
}

\date{February 2024}

\begin{document}

\maketitle

\begin{abstract}
The typical phases of Bayesian network (BN) structured development include specification of purpose and scope, structure development, parameterisation and validation. Structure development is typically focused on qualitative issues and parameterisation quantitative issues, however there are qualitative and quantitative issues that arise in both phases. A common step that occurs after the initial structure has been developed is to perform a rough parameterisation that only captures and illustrates the intended {\em qualitative} behaviour of the model. This is done prior to a more rigorous parameterisation, ensuring that the structure is fit for purpose, as well as supporting later development and validation. In our collective experience and in discussions with other modellers, this step is an important part of the development process, but is under-reported in the literature. Since the practice focuses on qualitative issues, despite being quantitative in nature, we call this step {\em qualitative parameterisation} and provide an outline of its role in the BN development process.
\end{abstract}

A Bayesian network (BN) is a directed acyclic graph (DAG), the nodes of which represent random variables, and the arcs of which represent conditional probabilistic dependencies~\citep{pearl1988}. A causal BN is one in which the directed arcs are to be interpreted as causal~\citep{pearl1995}. Development of a BN~\citep{korb&nicholson2010} model passes through several typical stages that include specifying the purpose and scope of the model, developing the structure (including selecting variables and specifying links, i.e. dependencies, between variables), parameterisation and validation. Structure development typically focuses on qualitative issues, such as the nature of variable definitions and how they are related to each other --- causally, if the BN is causal. Parameterisation typically focuses on quantitative issues, such as the specific probabilities of one variable conditional on another. However, there is another small step used in common practice when developing a BN that sits between structure development and full parameterisation. This common extra step involves developing a parameterisation focused solely on capturing the {\em qualitative behaviour} of the model. It is frequently used to iteratively improve the structure before a more rigorously quantitative parameterisation, and is also used as a prior, a guide and a check for that later quantitative parameterisation.

Here we define the {\em behaviour} of a BN model as the probability changes that occur in response to observations or (for a causal BN model) interventions. The {\em qualitative} behaviour of a BN model is the {\em way} in which the probabilities change. This includes the direction of change, constraints on the way probabilities can change, the relative size of changes and anything else that is not specifically quantitative. Qualitative behaviour is often testing during validation, but cannot be captured by the structure alone and is not often an explicit goal of the quantitative parameterisation. Our confidence and knowledge of the qualitative behaviour of a BN model is typically {\em significantly} higher than our confidence in the precise quantitative behaviour. In many kinds of decision making, it is also the qualitative behaviour that matters most. Indeed, some BN modelling software tools have been developed to focus exclusively on capturing qualitative behaviour~\citep{wellman1990,druzdzel2009}. We believe it is worth explicitly recognising this qualitative kind of parameterisation as an extra stage in the full, quantitative BN development process. {\em Qualitative parameterisation} is the new term we propose for this previously unnamed practice.\footnote{In previous work in which members of our group were involved~\citep{boneh_ontology_2010}, we have referred to a similar approach to support prototyping as `handcrafting'.} While this practice often goes unreported in the literature, our collective experience of over 20 years in building over 100 BNs, as well as discussions with many other modellers suggests this practice is extremely common. We describe what qualitative parameterisation is and the role it plays during BN development in more detail below.

\section{Bayesian network development}

\begin{figure}
    \centering
    \includegraphics[width=0.9\textwidth]{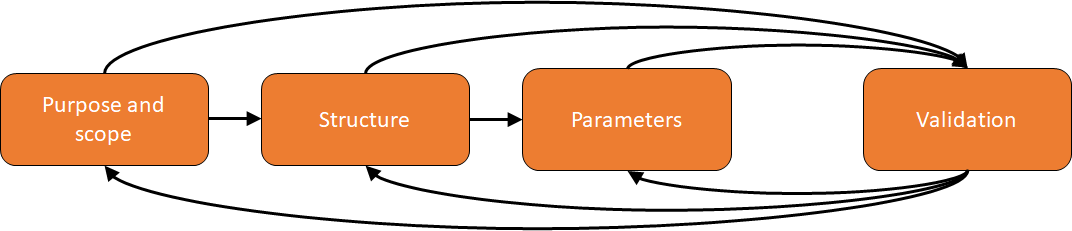}
    \caption{A simplified overview of the iterative BN development process}
    \label{fig:devprocess}
\end{figure}

The development of a BN typically proceeds iteratively through a number of common stages (Ch. 10,~\citealp{korb&nicholson2010}). These stages have much in common with the typical iterative software development process, however there are some unique considerations for BNs. Typically, the purpose of the BN is first defined (essentially, its functional requirements) and its scope settled. Development is then split into two stages that are specific to BNs, in that there is a structure development stage (i.e., development of the DAG), followed by a parameter development stage (typically, the conditional and unconditional probabilities). The final stage (not counting iterations) is the validation stage, in which the behaviour, accuracy and capabilities of the BN are tested to see if the BN fulfils its purpose and scope. See Figure~\ref{fig:devprocess}.

While the stages are iterative, they are also dependent. The purpose and scope for a BN primarily drives its structure and what it will include. Indeed, the purpose and scope are {\em required} for structure development. Similarly, the structure is {\em required} before the parameters can be provided. (And note that changes to any earlier stage typically necessitate changes at all later stages.) Validation is somewhat different. Not all the preceding stages are required before validation can be performed. For example, the purpose and scope can be validated before the structure is given (e.g., by ensuring it represents stakeholder requirements correctly), and the structure can be validated before the parameters are given. However, this is not always true. To validate that the scope is adequate for the purpose, it may be useful to make use of the structure; similarly, to validate that the structure is adequate to the purpose, it may be useful to use the parameters. 

The iterative nature of BN development means we often want to validate our BN before it is a complete BN. It may be incomplete due to missing structure or parameters. For incomplete structures, we can use temporary working definitions for variables, select substructures, or simplify intermediate structures so that we can proceed to the next stages. For incomplete parameters, things can be more difficult as there may not be any subsets or simplifications that can be easily made. If we still wish to validate the structure --- for example, to check that the variable definitions and arcs are able to capture the intended behaviour of the model --- we need to provide a parameterisation that at least supports the expected qualitative behaviour.

As part of the development process, we typically also want support for performing quantitative parameterisation. For example, if we're training the BN from data, we would like some way of dealing with an inadequate amount of data and some way of assessing how well the parameterisation based on data matches our expectations. Even when working with experts, we typically want some way to check that the set of parameters elicited from experts illustrate the qualitative behaviour we expect when they are put together in the complete BN. A qualitative parameterisation can provide these kinds of priors and checks against expectations.\footnote{Caution needs to be exercised when using a qualitative parameterisation as a prior, as the prior should really be on the qualitative relationships and not the precise numbers.}

Finally, our knowledge about the qualitative behaviour of a BN model is typically much higher than our knowledge of the precise quantitative behaviour of a BN model. For example, we may be extremely confident that taking paracetamol will substantially reduce a temperature, but not know precisely by how much for any given temperature. We can still use a BN model that captures this qualitative relationship to come to qualitative conclusions, which of course can also be used to validate the BN model, both its structure and future quantitative parameterisations.

\section{Qualitative and quantitative development}

Within the BN community, the parameters of the model are often considered part of the quantitative component of the model, with the remainder commonly considered to make up the qualitative component. In casual use, quantitative and qualitative are sometimes treated as synonyms for parameters and structure, respectively. This is largely because the only numerical (and hence quantitative) components of a BN are frequently the parameters, while the non-numerical (and hence qualitative) components are often fully covered by a specification of the structure. However, this is not always the case. For instance, when no explicit numerical parameters are provided, and only constraints on the type of relationship, the model as a whole may be called qualitative~\citep{wellman1990}. Similarly, each node often has local structure (such as an underlying decision tree or logit model structure) that cannot be determined from the BN's global structure~\citep{mascaro&woodberry2022,koller&friedman2009}. These aspects of the BN are also qualitative. Another common practice is to add constraints as child nodes of the parents being constrained~\citep{crowley+2007}. These constraint nodes are often deterministic and best considered entirely as part of the qualitative component of a BN, as they specify a quality or feature of the model (e.g., that exactly one of two nodes must be true). 

The convention of using the term `quantitative' just for parameters, and the term `qualitative' just for structure is also often appropriate because it matches the typical {\em provenance} of each type of information. For example, the parameters in the BN are often derived from measurements of some specific underlying value that is itself quantitative were it to be directly observable. When not derived from measurements, they may be estimated by experts, but the goal of the experts is to estimate the specific underlying quantitative value. In either case, the underlying value is quantitative. This is not always the case. For instance, parameters may be elicited using qualitative or semi-qualitative methods, such as by presenting verbal interpretations of probabilities~\citep{hope+2002,vandergaag+2012}, though the underlying goal is still a numerical estimate that is likely much more precise than the estimates from a qualitative parameterisation.

On the other hand, structure, and particularly causal structure, is aimed at capturing whether or not variables are related and the causal direction of that relationship (i.e., its quality and nature), regardless of the specific numbers that may determine the relationship. Typically, it is experts that provide their judgements on what may be related and how in the underlying process, and numbers are rarely involved. Even for machine learned causal structures, experts will help to judge and correct the causal structure in almost all cases.

But if we use machine learning to learn a structure --- for simplicity, a {\em non-causal} structure --- based purely on a dataset consisting of measurements, there is a sense in which the structure has been developed quantitatively. We might call such a process `quantitative structure development'. This is because the relationships that are discovered have a strongly (and in this case purely) quantitative origin. In particular, arcs are omitted wherever there is conditional independence --- i.e., where the conditional mutual information is 0. When it is non-zero, arcs are included. Indeed, this extends beyond BN structure, and applies to local structure too. For example, if a data set for a node encodes a Noisy OR, the trained probability table could be identified as fitting a Noisy OR structure well and converted into one. It is clear we can say that structure has been quantitatively developed.

We can apply the concept in the inverse to parameters. If parameters have been developed purely or mostly on the basis of the qualitative features that they give rise to, we can call this development process `qualitative parameter development' or just `qualitative parameterisation'.

\section{Qualitative parameterisation}


Qualitative parameterisation involves the selection of parameters --- that is, the parameters in a conditional probability table or other local structure such as a Noisy OR --- so that they represent the desired {\em local qualitative} behaviour and typically contribute to the desired {\em global qualitative} behaviour. Unlike more formal kinds of parameterisation, qualitative parameterisation does not aim to provide an accurate estimate of the parameter. Instead, it aims to represent the qualitative behaviour --- that is, the way in which probabilities change as evidence or interventions are entered into the BN or when moving between different sets of evidence. 

\begin{figure}
    \centering
    \includegraphics[width=0.6\textwidth]{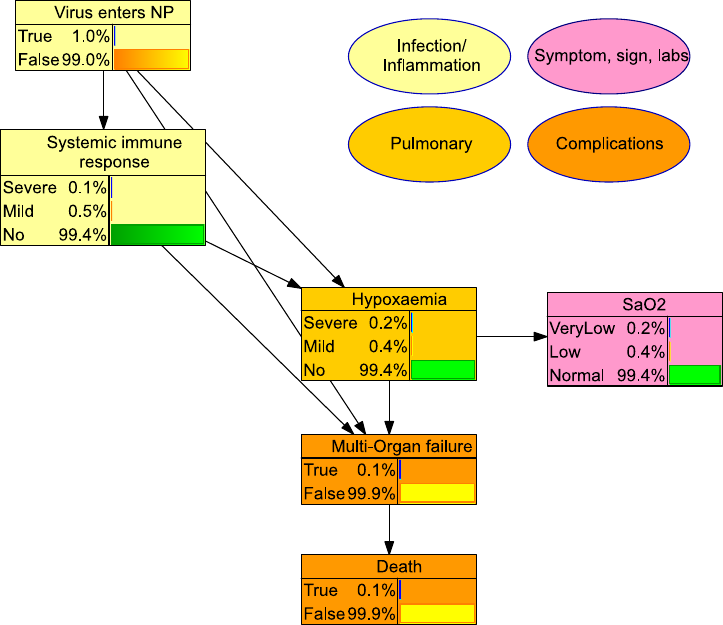}
    \caption{A simplified COVID-19 BN progression model that has been qualitatively parameterised. See \url{https://osf.io/bynr6/} for the full model.}
    \label{fig:qp-bn}
\end{figure}

Figure~\ref{fig:qp-bn} is a very small illustrative BN consisting of 6 nodes that has been qualitatively parameterised.\footnote{A GeNIe version of the example BN model can be found at \url{https://bayesian-intelligence.com/publications/qp/resp_simple_v1.0.xdsl}. The full qualitatively parameterised BN, called the Respiratory BN, can be found at \url{https://osf.io/bynr6/} and its structure is described in~\citep{mascaro+2023}.} The model represents simple progression of the disease, from entry into the nasopharynx, to development of an immune system response, the potentially resultant hypoxaemia (latent, but measurable from SaO2 measurements), followed by potential multi-organ failure and death.

In this case, the modellers have provided probabilities that aim to represent the qualitative information as understood at the time of parameterisation. For example, the model assumes a small chance that the virus will enter the nasopharynx. If it does so, it will result in a small chance of a severe immune response, a larger chance of a mild immune response, and the largest chance of no immune response. There would be some chance of hypoxaemia, which would drive a very indicative, but nonetheless noisy, SaO2 measurement (in this case, specified as approximately 95\% accurate). Both severe hypoxaemia and a severe systemic immune response will trigger a notable chance of multi-organ failure (around 28-33\%), and multi-organ failure, if it results, will lead to an 80\% chance of death.

There is no claim that these parameters are good estimates in themselves, but an effort was made to give them values that are plausible given what was known. More importantly, the modellers have tried to capture the major qualitative features of the relationships. For example, there is only a small chance of the virus entering the nasopharynx for some period of time (such as a day or a week); multi-organ failure has a significant chance of leading to death; hypoxaemia is likely very well measured by SaO2; severe immune responses are associated with severe hypoxaemia, and so forth.

Even if the specific parameter estimates are very crude, the quality of the estimates and how they relate to each other allows us to look at the overall response of the model to different sets of evidence. For example, entering evidence of a death shows us that there is a roughly 22\% chance that the virus had entered the nasopharynx. By contrast, observing no death does not affect the chance of a virus having entered the nasopharynx (still at approximately 1\%), and in fact does not affect many of the posterior probabilities in the model at all. Similarly, observing very low SaO2 suggests a good chance that a virus entered the nasopharynx, a reasonable chance of a severe immune response, and a non-negligible chance of death.

These are indeed the kinds of global qualitative behaviours that we would hope to see for this model, and they suggest that the {\em structure} is able to capture the right behaviour. This qualitatively parameterised BN does {\em not} suggest that we have any reason to believe the posterior probabilities would be at all accurate.

The qualitative parameterisation has further benefits for validating global behaviour. These are similar to the benefits that one gets from performing ordinary quantitative parameterisation. For example, it is possible that some of the arcs are not needed to capture the desired behaviour. This might be identified by the modeller as we did in the previous discussion, that is, by exploring different evidence scenarios, but it can also arise as the modeller performs the parameterisation itself. For example, the modeller may decide that there is no useful way to split hypoxaemia on the basis of {\em both} systemic immune response and virus entering the nasopharynx. They may realise that immune response may be enough on its own --- and this can then be verified with experts or with data. Similarly, as the modeller thinks through how to parameterise a node, they might identify a dependency that was previously missed --- e.g., perhaps death may also be dependent on the immune response. They may also come to a better understanding of the kinds of local structure that would be suitable --- e.g., perhaps the immune response might interact with hypoxaemia to make the chance of multi-organ failure worse; or perhaps they can be treated as completely independent causes. Hence, there can be many benefits from simply stepping through the process of parameterisation in this qualitative way, prior to a more rigorous quantitative parameterisation.

\vspace{1em}

\noindent Qualitative parameterisation is a frequent practice amongst BN modellers during the early stages of the BN development process, a practice which has until now gone without a name. The practice focuses on capturing the qualitative behaviour of a model, which allows one to validate the structure, explicitly identify and represent the qualitative relationships, as well as guide and check a later quantitative parameterisation. By recognising the practice, and explicitly understanding its benefits and limitations, we can not only make sure to use it in the development process where it is most helpful, but also to recognise situations where it is inappropriate and something more rigorous and faithful is required.

\bibliography{references}
\bibliographystyle{apalike}

\end{document}